\documentclass{article}
\usepackage{spconf}
\usepackage{amsmath}
\usepackage{multirow}
\usepackage{multicol}
\usepackage{booktabs}
\usepackage{graphicx}
\usepackage{subcaption}
\usepackage{xcolor}
\usepackage{wrapfig}
\usepackage{amssymb}
\usepackage{amsmath}
\usepackage{pifont}
\usepackage{capt-of}
\usepackage{colortbl}
\usepackage[para]{footmisc}
\usepackage{hyperref}

\def\prepara{{\vspace{5pt}}}
\newcommand\blue[1]{\textcolor{black}{#1}}
\title{Look, Listen and Recognise: Character-Aware Audio-Visual Subtitling}
\name{Bruno Korbar$^*$ \qquad Jaesung Huh$^*$ \qquad Andrew Zisserman\thanks{$*$ These authors contributed equally to this work. This research was funded by EPSRC Programme Grant VisualAI EP$\slash$T028572$\slash$1, and a Royal Society Research Professorship.}}
\address{Visual Geometry Group, Department of Engineering Science, University of Oxford, UK}
\begin{document}
\ninept

\makeatletter
\let\@oldmaketitle\@maketitle% Store \@maketitle
\renewcommand{\@maketitle}{\@oldmaketitle% Update \@maketitle to insert...
  \includegraphics[width=\linewidth]{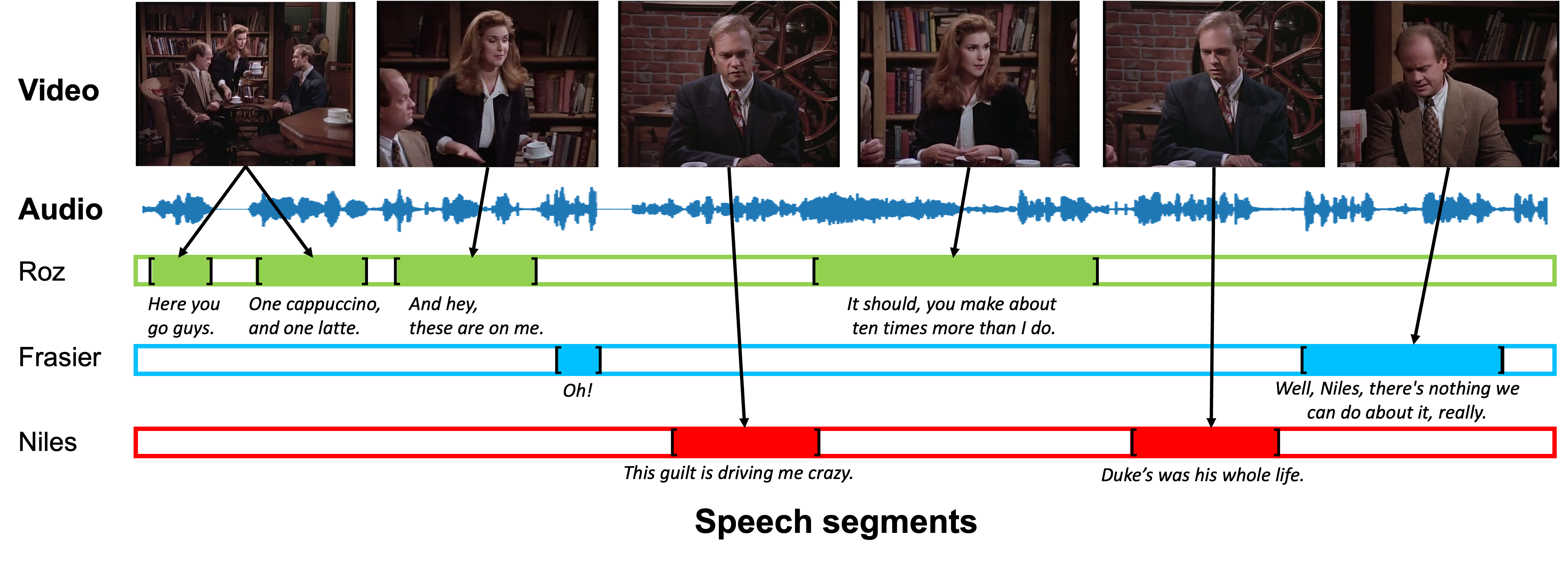} 
  \vspace{-0.3in}
  \captionof{figure}{Character-aware audio-visual subtitling. The generated data covers {\em what} is said, {\em when} it said, and by {\em whom} it is said.}
    \bigskip}% ... an image
\makeatother

\maketitle
% \twocolumn[{%
%   \renewcommand\twocolumn[1][]{#1}%
% \vspace{-0.4in}
%   \begin{center}
%     \centering
%     \captionsetup{type=figure}
%     \includegraphics[width=1\linewidth]{images/teaser_new.png}
%     \vspace{-0.3in}
%     \captionof{figure}{Character-aware audio-visual subtitling. The generated data covers {\em what} is said, {\em when} it said, and by {\em whom} it is said.}
%     \label{fig:teaser}
% \end{center}%
% }]

\begin{abstract}
The goal of this paper is automatic character-aware subtitle generation. Given a video and a minimal amount of metadata, we propose an audio-visual method that generates a full transcript of the dialogue, with precise speech timestamps, and the character speaking identified. The key idea is to first use audio-visual cues to select a set of high-precision audio exemplars for each character, and then use these exemplars to classify all speech segments by speaker identity. Notably, the method does not require face detection or tracking. We evaluate the method over a variety of TV sitcoms, including Seinfeld, Fraiser and Scrubs. We envision this system being useful for the automatic generation of subtitles to improve the accessibility of the vast amount of videos available on modern streaming services.
Project page : 
\url{https://www.robots.ox.ac.uk/~vgg/research/look-listen-recognise/}
\end{abstract}

\makeatletter{\renewcommand*{\@makefnmark}{}
% \footnotetext{$^*$Equal technical contribution. }\makeatother}

\vspace{-5pt}
\section{Introduction}
\vspace{-5pt}
\label{sec:intro}
With the rise of streaming platforms that allow watching videos ``on-demand'', more video content is made available to the general public and researchers than ever in history.
With more than 80\% of users of one such platform relying on subtitles~\cite{netflixuser}, automatic subtitle generation and captioning has become an important research topic in the community~\cite{radford2022robust, bain2022whisperx}. Unfortunately, many subtitles, whether automatically generated or not, do not comply with the standards for Subtitles for Deaf and Hard-of-hearing (SDH): namely, they do not include information about speaker identification, nor do they contain sound effects and music. 

\noindent In this paper, we take the next step towards automatic generation of SDH -- we aim to make the subtitles character-aware. 
Character-aware subtitles would also be of great benefit to researchers. 
They would allow for the automatic generation of large-scale video datasets, which could fuel the next generation of visual-language models capable of learning higher-level semantics from the paired data.

\noindent There has been a plethora of works using audio-visual networks for speech recognition~\cite{Afouras19, shi2022robust}, speaker diarisation~\cite{Chung20, ding2020self, xu2021ava} or character recognition~\cite{sharma2022audio,Everingham09,haurilet2016naming,Nagrani17b} which are subtasks of our main goal.
However, these works require additional processing for detecting and tracking faces.
% which are often quite heavy and time-consuming during inference.
We present a simpler method that does not require face detection or tracking and uses only off-the-shelf deep neural network models and the cast list for each episode.

% The goal of this effort is not too disimilar that of the VoxCeleb~\cite{Chung18b} dataset, which requires heavy pre-processing including face recognition and tracking. We strive to achieve our goal with minimal required metadata: only the knowledge of the cast and 1-5 images of each cast member, which can be gathered automatically.

\noindent We make the following four contributions: (i) we propose a new task, character-aware audio-visual subtitling, which aims to generate the \textit{what}, \textit{when} and by \textit{whom} for subtitles, with minimal required metadata. % \BK{(I think this is still ok, bc we specify audio-visual; bazinga is audio only?)} \az{ok}
(ii) we develop an automatic pipeline for this task that does not require face detection or tracking (Section~\ref{sec:method}); (iii) we curate an evaluation dataset that includes subtitles labelled with characters individually for three different sitcom series: Fraiser, Scrubs and Seinfeld (Section~\ref{sec:dataset}); and (iv) we assess the method on the evaluation dataset and report the performance (Section~\ref{sec:results}).

\vspace{-5pt}
\subsection{Related work}
\vspace{-5pt}
\noindent\textbf{Labelling people in videos.} is a well studied topic in computer vision~\cite{Everingham09,haurilet2016naming,Nagrani17b}. 
\blue{Often, the availability of various levels of prior information is required such as scripts~\cite{Everingham09}, clean images for actor-level supervision~\cite{Nagrani17b}, or ground truth subtitles with correct timestamps~\cite{mocanu2019enhancing,akahori2017dynamic}.} 
\cite{Brown21} relaxes the need for cleaned data and makes their method scalable by gathering a large amount of data via automated image search to obtain the corroborative evidence they use for supervision. 
Like~\cite{Brown21}, our model retrieves the necessary information via search engines, however, it does not pre-process video frames, save for the transformations required by a neural network.

\prepara\noindent\textbf{Audio-only speaker diarisation.}
 Speaker diarisation is the task of identifying  ``who spoke when'' from a given audio file with human speech.
 There are two branches of works in this area: (i) using existing Voice Activity Detection (VAD) and a speaker model together with clustering~\cite{wang2018speaker, zhang2019fully, kwon2021look} and (ii) using an end-to-end model which goes from the VAD to assigning speakers~\cite{fujita2019end, horiguchi2020end}.
Both of them suffer when the number of speakers is large such as in TV shows or dramas.
Furthermore, the current state-of-the-art speaker recognition models assume that the input is long ($>$ 2 sec), while most of the speeches in TV shows are relatively short including exclamations, which leads to the degradation of speaker clustering performance.
In this paper, we include the active speaker detection model and person-identification model, which are strong in short videos, to identify the character.

 \prepara\noindent\textbf{Audio-visual speaker diarisation.}
In the last few years, efforts were made to improve the performance of diarisation by borrowing the power of face recognition models or lipsync models, which are closely related to human speech~\cite{Chung20, xu2021ava, chung2019said}.
\cite{Chung20} utilises audio-visual active speaker detection model and speech enhancement models, but mostly in celebrity interviews or news segment where the length of speeches are generally short.
\cite{xu2021ava} introduces an Audio-Visual Relation Network (AVR-Net) that leverages the cross-modal correlation to recognise the speaker’s identity.
% \az{not clear what this network does} modality mask based on the presence of the face per each video segment.
%
Our approach is different from these works in two ways: (i) we do not use any face detection or tracking; and (ii) we introduce character-aware audio-visual subtitling that builds the character bank within each video and figures out not only the speaker clusters but the speakers' \textit{identity} for each utterances and the speech content.

\prepara\noindent\textbf{Datasets.}
The Bazinga!\ dataset~\cite{lerner2022bazinga} also provides subtitles labelled
with characters for a large number of TV series. However, it is an audio only dataset, and consequently is not directly suitable for applying the audio-visual approach we develop.

\vspace{-10pt}
\section{Method}
\label{sec:method}
\vspace{-5pt}

\begin{figure}
\centering
\includegraphics[width=\linewidth]{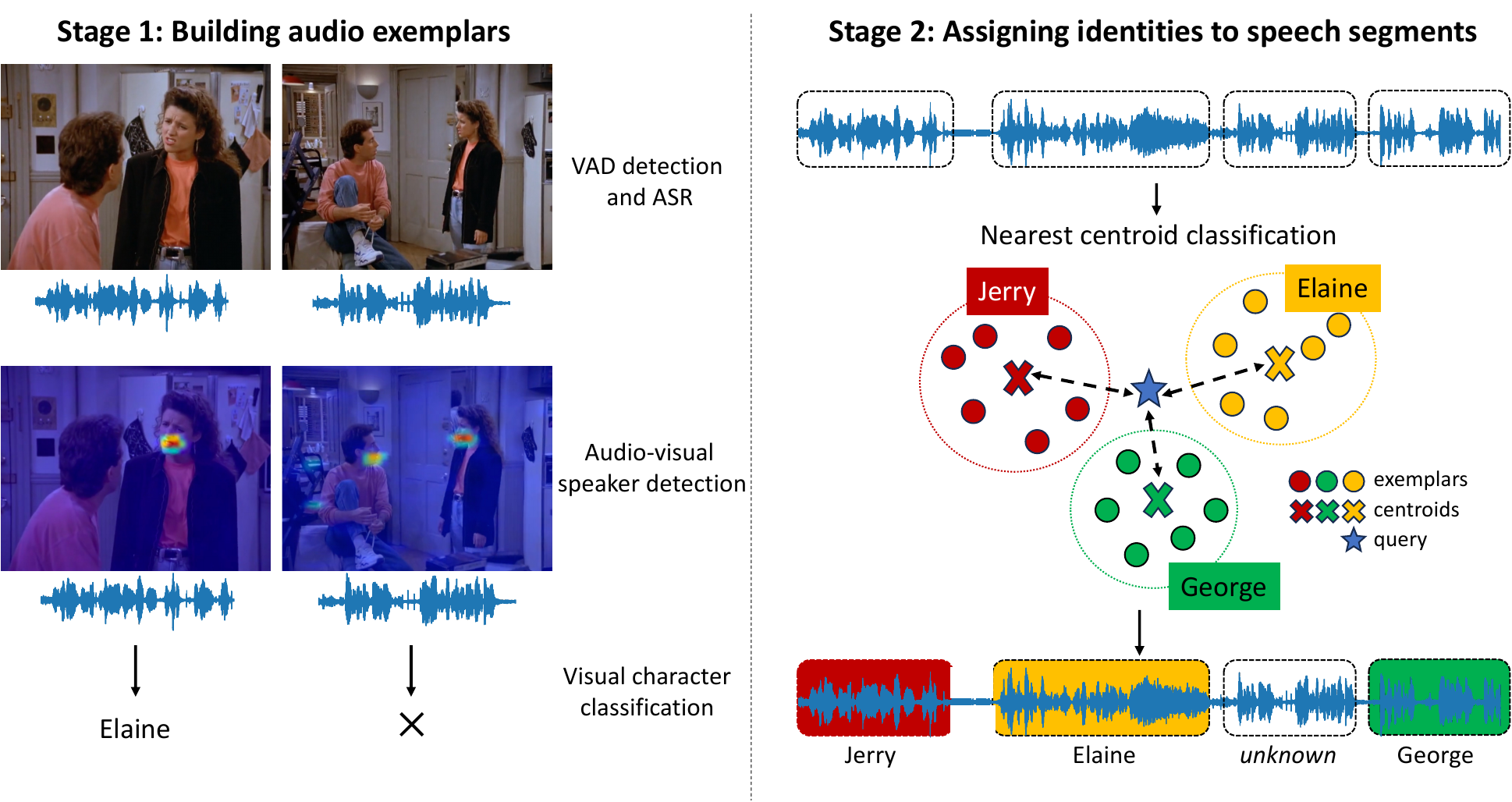}
\vspace{-15pt}
\caption{\blue{Overview of our method. We first build a database of audio exemplars for each character by filtering speech segments until only a high precision set remains (left). Each speech segment is then assigned to a character by comparing its voice embedding to the exemplar embeddings (right).}}
\vspace{-15pt}
\label{fig:methodoverview}
\end{figure}

This section explains our approach to creating subtitles for the video and attributing speakers to each speech segment.
% In order to effectively subtitle the video, we first detect speech segments and recognise what is being spoken in them. But assigning identities to a voice that you've never heard before poses a significant challenge. For example, the only way we can recognise if `George' is speaking off-camera (or where there are multiple characters present, any of who could be speaking) is if we know what he sounds like. By finding examples where George is clearly the only person speaking in the video, we can be confident we have a good example of his voice. We can then match the voice from these examples that we are certain are correct to other speech utterances in the video. 
Our method consists of two distinct stages. 
First, we detect speech segments from the video, recognise the spoken words, and process the data to create a database of what we refer to as \textit{speech exemplars} -- sample video clips where a speaker is clearly audible, visible and identifiable. In the second stage, the speech exemplars for each character are used to assign the identities to {\em all} speech segments. 
% Each step of the exemplar creation is outlined in Section~\ref{sec:methodexemplars}, and the process of using the exemplars for annotation of the whole video is explained in Section~\ref{sec:methodannotation}. The method is illustrated in Figure~\ref{fig:methodoverview}. 

% For illustrative purposes, we discuss each step for the Seinfeld TV show, but an analogous process is used on the other TV series in the dataset.}

% We detect voice-activity regions (what we refer to as VAD segments) by using differnt VAD models, such as WhisperX~\cite{bain2022whisperx} or pyannote~\cite{Bredin2020} and also filter out laugh track using off-the-shelf laughter detector. 
% Then, for each of the VAD segments, we verify the visibility of the speaker by detecting audible objects and confirming the audible object is clearly visible in the frame~\cite{Afouras20b}. 
% Finally, we classify the visible speakers using CLIP-PAD~\cite{Korbar22} model. If we can confirm that for a given segment, only one speaker is audible, visible, and recognisable, we deem that VAD segment an ``exemplar''. We later use these exemplars to classify all detected speech regions. 

In order to label the characters we require the following metadata for each episode: (i) the names of the characters in the show; and (ii) for each character 1--10 sample images of the actor and their names that we can use as visual examples. \blue{This metadata can be obtained automatically from online database of movies or TV series~\cite{imdbcom}.}

\vspace{-5pt}
\subsection{Stage 1: building audio exemplars}
\vspace{-5pt}
\label{subsec:build_exemplars}
% \BK{see edit below} \az{This has implementation details but not what is done and why. First give an overview of the steps explaining what is done and why for each. Use the Seinfeld numbers/table as a running example. After that give the implementation details below.}
% \JH{What is input and what is output, also each step.}

The goal of stage 1 is to create a database of character voices. We take multiple episodes of a TV series, and obtain a set of speech segments for each character.

\noindent In order to do this, we first split videos into speech segments, and transcribe them.
For each segment we determine if only one speaker is visible and is speaking -- a crucial step because it allows us to be confident that the speech segment corresponds to the face in the frame.
\blue{We collect a set of speech segments for each character that we can confidently recognise from their face, and then filter the samples in each set to remove potential label noise using voice embeddings.}
% (for example in the rare occurrence that George is visible , but Kramer is speaking off-camera).
%
We end up with a set of speech segments for each character that are  recognised with high  precision, and refer to these as \textit{speech exemplars}. 
The building of these exemplars is illustrated in Figure~\ref{fig:methodoverview}, and
we give details of each sub-step below.

\prepara\noindent\textbf{1.\ VAD detection and Automatic Speech Recognition (ASR).}
In this stage, we take an entire video and split it into segments where speech is detected and recognised. 
\blue{We first detect the voice regions across the entire dataset and determine the spoken content of each segment.
We do this with a language-guided VAD model. 
We apply the WhisperX~\cite{bain2022whisperx} model on the audio stream of our dataset which detects the speech regions with word-level timestamps.
% These word-level segments are extremely short ($<$0.5 sec) and require a separate algorithm to merge them.
We concatenate the generated words to obtain the entire transcription per video, then use a sentence tokenizer to separate them by sentences.
Assuming each sentence is spoken by a single speaker, we use the start and end times of the sentences as our unit of speech segments.}
% We also run a voice segmentation model~\cite{Bredin2020} on our dataset and remove the false-positive voice segments generated from WhisperX by rejecting the voice segments that are not detected by both models.
%

\noindent We also find that most TV shows contain laughter tracks (audience laughter) which are voice regions but are not of interest to this work.
Thus, we run a pretrained laughter detector~\cite{gillick2021robust} for each of the remaining voice segments and remove the ones from the candidates of exemplars if laughter is detected.
After this step, we know precisely when characters in the show are speaking and what they are saying. We don't yet know \textit{who} is saying what. 

% Note that we can also run the audio-only diarisation model on top of these segments and use them for our final character predictions per each segment.
% We extract speaker representation using pretrained ECAPA-TDNN~\cite{desplanques2020ecapa} per each segment.
% Segments are then clustered using Agglomerative Hierachical Clustering (AHC).

\prepara\noindent\textbf{2.\ Audio-visual speaker detection.}
% \JH{Input and output}
\blue{
The goal of this stage is to take speech segments from the previous stage and select only those with a single visible speaker. This will produce a subset of speech segments where we can recognise the speaker.
To achieve this, we localise the speaker with an audio-visual synchronisation model~\cite{Afouras20b} which produces a spatial location of the audible objects and has been shown to detect speakers well. 
In practice, it generates an audio-guided heatmap over each video frame. 
We average the heatmaps over the length of each speech segment to avoid unnecessary noise and detect peaks in the heatmap through a combination of maximum filtering and non-maximum suppression. 
Example heatmap outputs can be seen in Figure~\ref{fig:methodoverview}.
When a single peak is visible throughout the video clip, we can assume that only one speaker is present. If there are no detected peaks, or there are multiple ones, we discard that speech segment from the candidates of exemplars.}

\prepara\noindent\textbf{3.\ Visual character classification.}
In this step a character name is assigned to each of the single-speaker speech segments from the previous step where possible. This leaves us with a further reduced set of speech segments, each having a character name associated with it. 
Character classification is the only step in our annotation process that external data is used.  Specifically, the 1--10 sample images of each actor are used to form a visual embedding of that character.
% We describe an automatic process for gathering this information in the supplementary material. 
% We take the subset of single-speaker speech segments from the previous step and assign character names to them. This leaves us with a further reduced set of speech segments, each having a character name associated with it. 
%
Our classification model~\cite{Korbar22} compares a visual embedding of the frames of a speech segment to a combination of actor visual embedding and actor name (details are given below). We select the best match or discard the clips which cannot be classified with a high degree of confidence. Note, (i) the comparison is at the frame level, no face detector or cropping is required for this visual recognition; (ii)  we compute visual embeddings for all characters in a given season, but only consider ones present in that episode at inference time. 

\prepara\noindent\textbf{4.\ Audio filtering.}
Finally, we group the labelled speech segments from the previous stage by character, and for each character  we filter their {\em voice} samples to remove potential noise from the groupings as follows:
 we compute voice embeddings for each sample, and consider that a sample is positive for a given character if its 5 nearest neighbours are labelled as the same character. Note that for characters where the number of samples $n$ is smaller than 5, we keep all the samples in our database. 
This gives us the final exemplar set for a given TV series and hopefully leaves us knowing what each character sounds like.

\vspace{-5pt}
\subsection{Stage 2: Assigning characters to speech segments} 
\vspace{-5pt}
\label{subsec:methodannotation}
\blue{The aim of this stage is to assign a character name to each of the detected audio segments that we are confident of, regardless of whether a speaker is visible or not.
On a high-level, we achieve this by comparing the distance between each speech segment and the audio exemplars for each character.
We do not assign an identity if the minimum distance is above a certain threshold.}

\noindent\blue{Specifically, for each character we compute the mean of exemplar embeddings and use it as a centroid representation for that character.
To classify speech segments, we embed them with the same model used to generate the exemplar embeddings, and measure distances to class centroids. 
The segment is assigned to the speaker corresponding to the nearest centroid.
However, if the minimum distance between the segment embedding and each centroid is bigger than a threshold $d$, then that segment is classified as ``unknown''. 
This covers uncertainty and also the cases where we don't have exemplars.}
\vspace{-5pt}
 \subsection{Implementation details} 
 \vspace{-5pt}
 \blue{
 We detect speech and perform ASR with an off-the-shelf WhisperX~\cite{bain2022whisperx} model, and the sentences are tokenized with NLTK~\cite{bird2006nltk} tokenizer. 
 We use the laughter detector by \cite{gillick2021robust} with a detection threshold of 0.8.
 All voice embeddings are encoded with ECAPA-TDNN~\cite{desplanques2020ecapa}, which is pretrained with VoxCeleb~\cite{Nagrani19}.
% Segments shorter than 0.1 seconds are zero-padded before putting into the speaker model.
%
For discovery of speaking faces, we use a pretrained LWTNet~\cite{Afouras20b}. For each generated heatmap we detect 4 peaks, and consider each a positive if it's larger than $\tau_{\text{det}}=0.7$.
For actor face classification, we use the CLIP-PAD model~\cite{Korbar22} pretrained on VGGFace and VGGFace2~\cite{Parkhi15}. Actor text-image embeddings are formed as \texttt{"An image of <TKN> Name Surname"} where \texttt{<TKN>} is an average representation of query images computed using a face-embedding network, as in \cite{Korbar22}. 
To classify the actors in the scene, we measure the cosine similarity between the visual embedding of the frames and the text-image embedding and choose the ones with highest similarity score where the score is over threshold $\tau_{\text{rec}}=0.85$ as positives. 
All hyper-parameters are determined via grid search on the three validation episodes, and kept fixed otherwise.}
%when applied to the test episodes of the dataset.}

% \BK{Version 1} Then, for each character, we train a one-vs-rest SVM classifier and use them to classify other speech segments. We report the results in Section~\ref{sec:results}.

% \BK{Version 2} Then, to classify each speech segment we embed it with the same model used to generate the exemplar embeddings, and then use a kNN classifier with majority vote to obtain final prediction. We report the results in Section~\ref{sec:results}.

\vspace{-5pt}
\section{Evaluation Dataset}
\vspace{-5pt}
\label{sec:dataset}
In this section, we describe a semi-automatic annotation pipeline used to generate the ground truth character names, timestamps and subtitles for speech segments.
The goal is to annotate the identities for all subtitles with accurate time intervals in the video.

% We use the transcripts from online sources and speech recognition results with word-level timestamps to produce the initial results followed by manual verification.

\vspace{-10pt}
\subsection{Annotation procedure}
\vspace{-5pt}
\label{subsec:annotation}
\blue{
The dataset collection process consists of two stages: (i) automatic initial annotations by aligning a transcript with timed subtitles; and (ii) human annotators reviewing and further refining these annotations.
Note that our dataset differs from other speaker diarisation datasets since we are also interested in the \textit{identity} of each speaker and speech transcriptions.}

\prepara\noindent\textbf{Aligning transcripts and timestamps.} \blue{To associate character names with corresponding temporal timestamps, we leverage two readily accessible source of textual video annotation: original transcripts and subtitles with word-level timestamps.
Transcripts are obtained from multiple online sources~\cite{frasiertranscripts, seinfeldtranscripts, scrubstranscripts}.
They include spoken lines and information about who is speaking. However, they do not provide any timing information beyond the order in which the lines are spoken. 
We use WhisperX~\cite{bain2022whisperx} to obtain the timed subtitles.
We find this suitable since its transcription and timestamps are highly accurate, whereas the timestamps in subtitles from other online sources often do not align with the actual speech in the video.
To align the original transcripts and timed subtitles, we employ the approach from~\cite{everingham2006hello}.
We use Dynamic Time Warping (DTW) to obtain the word-level alignment between the transcript and timed subtitles to associate the speaker with each of these words. Please refer to the original paper for the detailed process.}

\prepara\noindent\textbf{Manual correction.}
The output of the automatic pipeline is prone to several errors such as (i) a mismatch between the text of the transcript and WhisperX's transcription results; and (ii) mispredicted timestamps.
We correct any errors in timestamps and character names manually using the  VIA Video Annotator~\cite{dutta2019vgg}.

\vspace{-5pt}
\subsection{Dataset statistics}
\vspace{-5pt}
\label{subsec:dataset_stats}

% \begin{figure}[t]
%     \centering
%     \includegraphics[width=0.8\linewidth]{images/segment_length.png}
%     \caption{The distribution of utterance lengths over three datasets.}
%     \label{fig:length_dist}
% \end{figure}

Three TV series datasets are used to evaluate our method. 
\blue{We annotate the first six episodes of Season 2 of \textbf{Frasier}, Season 2 of \textbf{Scrubs} and Season 3 of \textbf{Seinfeld}.
We utilise the sixth episode in each season as our validation set, while the remaining episodes serve as our test set.
The detailed statistics are shown in Table~\ref{tbl:dataset_stats}.}
% Figure~\ref{fig:length_dist} shows the distribution of utterance lengths over three datasets.
% The video in this dataset mostly consists of short speech segments.
% Overall, 62\% of utterances are less than 2 seconds which are not suitable as inputs for existing speaker models, normally trained only with long utterances ($>$ 2 sec).

\begin{table}[t]
  \centering
  \footnotesize
    \caption{\blue{Evaluation dataset statistics. \textbf{\# episode}: number of episodes, \textbf{duration}: total duration of the dataset, \textbf{\#IDs}: total number of characters, \textbf{speech \%}: percentage of video time that is speech and \textbf{\# spks}: min / mean / max of number of speakers per video.}}
    \vspace{-5pt}
\label{tbl:dataset_stats}
\resizebox{1\linewidth}{!}{
\begin{tabular}{cclcccc}
\toprule
\textbf{Dataset} & \multicolumn{2}{c}{\textbf{\# episode}} & \textbf{duration} & \textbf{\# IDs} & \textbf{speech \%} & \textbf{\# spks} \\
\midrule
\textbf{Seinfeld}         & \multicolumn{2}{c}{6}                   & 2h 09m        & 36        & 60.6              & 6 / 9.2 / 12       \\
\textbf{Frasier}          & \multicolumn{2}{c}{6}                   & 2h 11m        & 29         & 59.5               & 6 / 9.2 / 12                   \\
\textbf{Scrubs}           & \multicolumn{2}{c}{6}                   & 2h 02m        & 48        & 67.9               & 13 / 15.7 / 18                   \\
\bottomrule
\end{tabular}}
\vspace{-15pt}
\end{table}

\section{Results} 
\label{sec:results}
\vspace{-5pt}
\blue{This section provides a detailed analysis of Stage 1 and 2, followed by the overall result on our test set.}

\vspace{-5pt}
\subsection{Detailed analysis of Stage 1 and 2}
\label{subsec:detailed_analysis}
\vspace{-5pt}

\prepara\noindent\textbf{Performance evaluation of Stage 1.}
\blue{
We evaluate the yield and classification accuracy of the speech exemplars on the five episodes of Seinfeld in our test set.
In Table~\ref{tbl:yield}, it can be seen that \textbf{19.3\%} of voice activity segments can be considered as exemplars. 
We also evaluate the performance quantitatively by manually inspecting the exemplars. The results, shown in Table~\ref{tbl:charperformance}, demonstrate that the accuracy of Stage 1 is almost perfect, being \textbf{100\%} correct for most characters. 
% We have on total \textbf{329} for the four main characters in Seinfeld, and  \textbf{78} for minor characters. 
There are 11 characters for which we have no exemplars in the 5 episodes of Seinfeld. They cover only 1.8\% of speech segments -- most of them speak less than five sentences in the episodes.}

\begin{table}[t]
\footnotesize
\centering
\caption{Exemplar yield after steps in Stage 1 (on Seinfeld).}
\label{tbl:yield}
\vspace{-5pt}
\begin{tabular}{@{}lcc@{}}
    \toprule
    Step                      & \# of exemplars & \% of total \\ \midrule
    VAD detection             &     2107            & 100.0 \\
    Audio-visual speaker detection &     1271            &  60.3 \\
    Visual character classification      &      806           &  38.3 \\
    Audio filtering           &      407           &     19.3        \\ \bottomrule
\vspace{-10pt}
\end{tabular}
\end{table}

\begin{table}[!t]
\footnotesize
\centering
\caption{Exemplar recognition performance for named characters in Stage 1 in Seinfeld. ‘others’ is a group of 21 characters, all named correctly.}
\label{tbl:charperformance}
\vspace{-5pt}
\begin{tabular}{@{}lccc@{}}
    \toprule
    Char. name             & \# exemplars & \# correct & Acc (\%) \\ \midrule
    Total                  &   407           &   406         &   99.8       \\
    Jerry                  & 273          & 272        & 99.6     \\
    Elaine                 & 30           & 30         & 100      \\
    Kramer                 & 12           & 12         & 100      \\
    George                 & 14           & 14         & 100      \\
    \textit{others}       & 78           & 78         & 100      \\ \bottomrule
\end{tabular}
\vspace{-5pt}
\end{table}

\begin{figure}[!t]
    \centering
    \includegraphics[width=\linewidth]{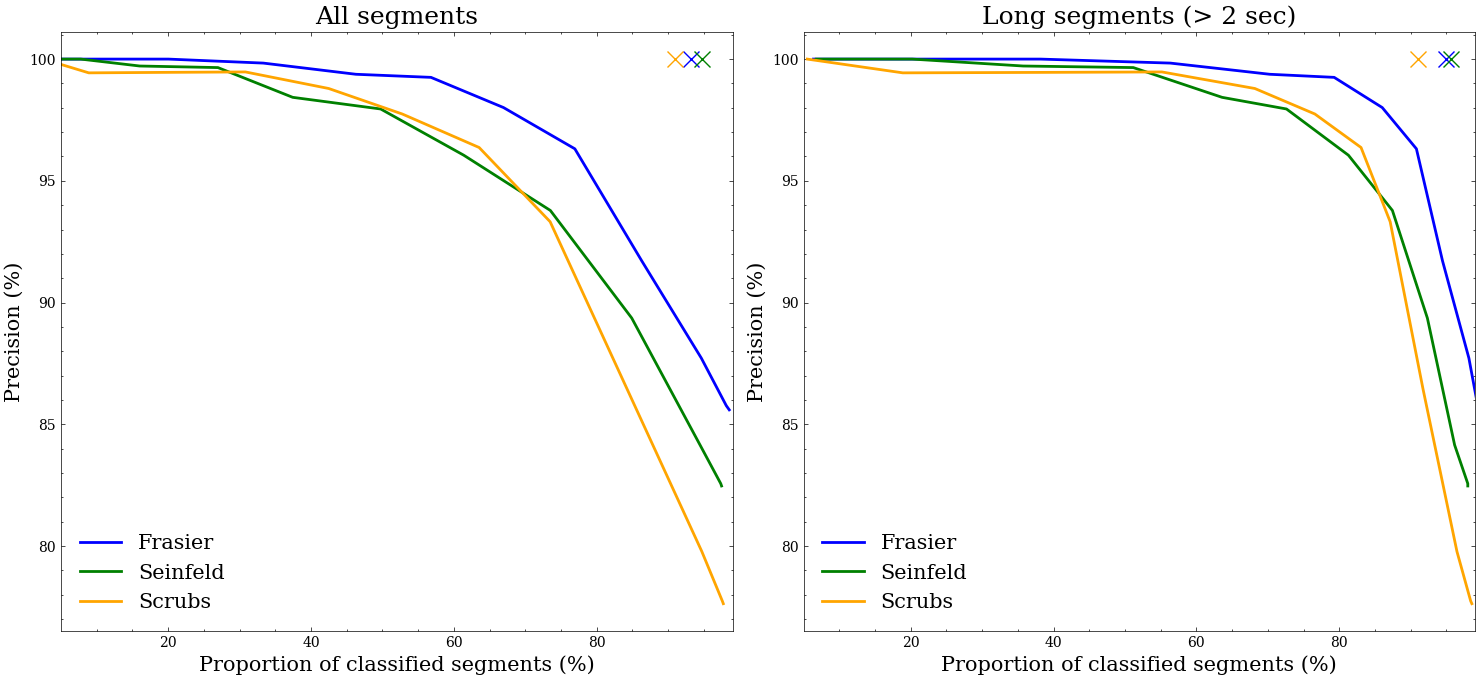}
    \vspace*{-10pt}
    \caption{\blue{Stage 2 Precision-POCS Curves for the test set of the three TV series, obtained by varying the threshold $d$ (for classification as ``unknown''). The left figure shows the performance using all detected speech segments. The right figure shows the performance only for the long segments ($>$ 2 sec). We also show the oracle points (‘x’ in each graph) for each TV series. The oracle point is where all segments for which there are character exemplars are correctly classified, and other segments are classified as ``unknown''.}}
    \label{fig:precision}
    \vspace{-12pt}
\end{figure}
\prepara\noindent\textbf{Performance evaluation of Stage 2.}
\blue{We demonstrate the trade-off between the Proportion of Classified Segments (POCS) and 
 overall precision by varying the threshold $d$ used in the nearest centroid voice classification to assign speech segments as ``unknown''. True positives are the segments that overlap with the ground truth segments and the character is correctly identified.
Figure~\ref{fig:precision} shows the result.
It can be seen that precision decreases as we classify more segments.
Also, long segments show higher precision in all three TV series at any given POCS, which shows that the speaker model produces better representations for longer segments.}

% \prepara\noindent\textbf{Cost of processing} Our visual speaker detection and recognition from stage 1 ** why are we restricting to stage 1 here? ** takes about 15 minutes on a single P40 GPU ** what is this measuring? all five episodes? ** , not accounting for data copying. Compared to over 40 minutes with higher CPU by~\cite{Brown21}. \JH{I need to check this with Bruno. Ask Bruno.} 

\vspace{-5pt}
\subsection{Overall performance on the test set}
\label{subsec:overall_results}
\vspace{-5pt}

% We evaluate our method on a manually annotated dataset containing five episodes of Seinfeld, Frasier and Scrubs. 
\prepara\noindent\textbf{Performance measures.} In addition to the traditional diarisation metric of Diarisation Error Rate (DER), we report the overall character recognition accuracy as well as the average of the per-character precision and recall metrics for the characters of each show. 
We use a 0.25-second collar to calculate DER.
Accuracy is calculated for the segments that overlap with one of the ground truth segments.

The results are given in Table~\ref{tbl:final}.
We can see that the model performs best on Frasier and worst on Scrubs in all metrics.
This is due to the difference in size of the casts in each dataset.
Scrubs has more characters than Frasier (48 $>$ 29) for a similar total duration (see Table~\ref{tbl:dataset_stats}).
Thus, Scrubs provides more potential assignments for each segment, making identification more challenging.

\noindent We also report the diarisation performance with and without the overlapping speech in Table~\ref{tbl:final}. 
% Note, Seinfeld and Frasier do not show much difference in DER between considering overlapping speech or not, meaning that there is not much overlapping speech within these two shows.}
The difference in DER for these two categories is small in Seinfeld and Frasier,  meaning that there is not much overlapping speech within these two shows.

\prepara\noindent\textbf{Speech transcription performance.}
Our method uses the WhisperX ASR model which also produces the speech transcription results.
We compare the performance with the state-of-the-art models in Table~\ref{tbl:wer}.
Word Error Rate (WER) is computed after applying the Whisper text normaliser to both ground truth and predictions which can be found in the original paper~\cite{radford2022robust}.
We see that WhisperX outperforms both Wav2vec2.0 and Whisper.
This is because the VAD Cut \& Merge preprocessing reduces the hallucination of Whisper, which is also mentioned in the original paper~\cite{bain2022whisperx}.

\begin{table}[t]
\centering
\footnotesize
\caption{\blue{Performance on the test set. We report the Diarisation Error Rate both with and without consideration of the overlapping regions, \textbf{DER(O)} and \textbf{DER} respectively. \textbf{Acc} denotes a character recognition accuracy for the segments that overlap with the groundtruth. \textbf{Ppc} and \textbf{Rpc} are the average per-character precision and recall, respectively.}}
\label{tbl:final}
\vspace{-5pt}
\resizebox{0.9\linewidth}{!}{%
\begin{tabular}{lccccccccccccc}
\toprule
Showname & DER$\downarrow$ & DER(O)$\downarrow$ & Acc$\uparrow$ & Ppc$\uparrow$ & Rpc$\uparrow$ \\ \midrule
Seinfeld & 29.6 & 29.7         & 81.2         & 0.922        & 0.841        \\
Frasier  & \textbf{23.8} & \textbf{24.3}& \textbf{83.1}& \textbf{0.933}& \textbf{0.888}  \\
Scrubs   & 32.6 & 36.4         & 76.1         & 0.883        & 0.853        \\ \bottomrule
\end{tabular}}
\vspace{-5pt}
\end{table}

\begin{table}[t]
\centering
\footnotesize
\caption{Word Error rate (WER) (\%) on each dataset.}
\vspace{-5pt}
\label{tbl:wer}
\begin{tabular}{llccc}
\hline
Model & Version &Seinfeld & Frasier & Scrubs \\ \hline
Wav2Vec2.0~\cite{baevski2020wav2vec} & \texttt{ASR\_BASE\_960H} & 45.0 & 36.9 & 36.3 \\
Whisper~\cite{radford2022robust} & \texttt{medium.en} & 13.2 & 13.5 & 10.6 \\
WhisperX~\cite{bain2022whisperx} & \texttt{medium.en} & \textbf{11.8} & \textbf{11.2} &  \textbf{9.2} \\ \hline
\end{tabular}
\vspace{-5pt}
\end{table}

\prepara\noindent\textbf{Qualitative example.}
\blue{We show a qualitative example of our results in Figure~\ref{fig:qual}. 
As can be seen, our method assigns the character for each speech segment, as well as timestamps and the transcription.}

\begin{figure}[!t]
    \centering
    \includegraphics[width=\linewidth]{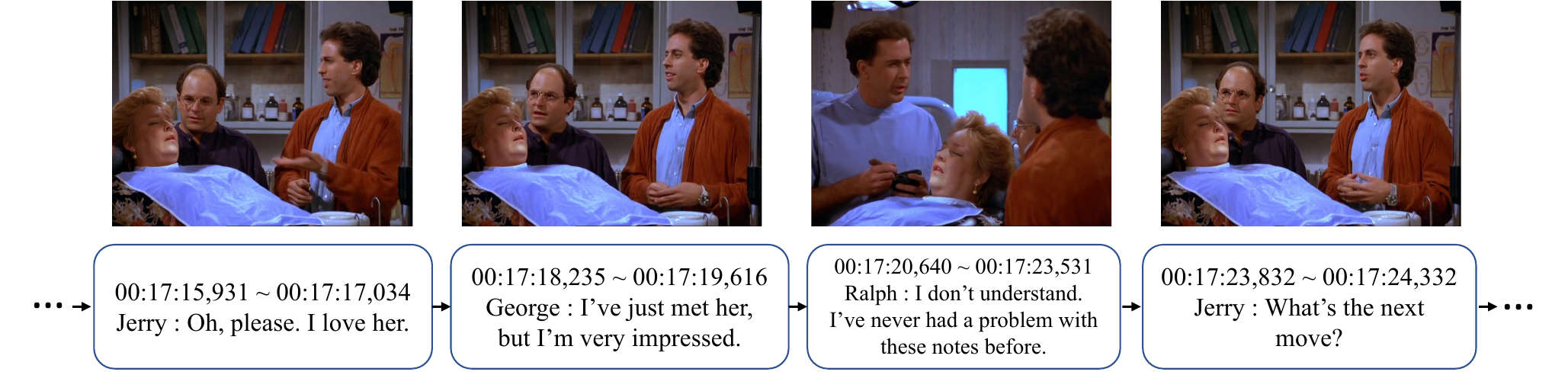}
    \vspace{-15pt}
    % \caption{Qualitative example. Our method produces the speech segments with timestamps and transcription, and assigns the character who spoke it.}
   \caption{Qualitative example. Our method produces the speech segments with timestamps, and assigns the character who spoke it.}
    \vspace{-15pt}
    \label{fig:qual}
\end{figure}

\vspace{-10pt}
\section{Conclusions}
\vspace{-5pt}
\blue{
In this work, we show promising first steps towards a model for character-aware subtitling, which we hope will be beneficial for improving accessibility, and facilitating further research in video understanding. 
Our method is not perfect, however. 
Our recognition efforts fail on short segments such as exclamations and also do not deal with overlapping speech -- though the latter does not appear to be a serious limitation in practice.
Furthermore, to generate the true SDH subtitles, we would need to classify and categorise every sound, not just speech -- something our model is not yet capable of.}
% We keep these considerations for future work. 

\bibliographystyle{IEEEbib}
\bibliography{shortstrings,vgg_local,vgg_other,egbib}

\end{document}